\documentclass[sigconf]{acmart}

\newcommand{\System}[0]{SEA}
\newcommand{\Cluster}[0]{ClusterEA}

\newcommand{\HitOne}[0]{H@$1$}
\newcommand{\HitTen}[0]{H@$10$}
\newcommand{\MRR}[0]{MRR}
\usepackage{listings} 
\definecolor{codegreen}{rgb}{0,0.6,0}
\definecolor{codegray}{rgb}{0.5,0.5,0.5}
\definecolor{codepurple}{rgb}{0.58,0,0.82}
\definecolor{backcolour}{rgb}{0.95,0.95,0.92}
\usepackage{subfigure}

\lstdefinestyle{mystyle}{
    backgroundcolor=\color{backcolour},   
    commentstyle=\color{codegreen},
    keywordstyle=\color{magenta},
    numberstyle=\tiny\color{codegray},
    stringstyle=\color{codepurple},
    basicstyle=\ttfamily\footnotesize,
    breakatwhitespace=false,         
    breaklines=true,                 
    captionpos=b,                    
    keepspaces=true,                 
    numbers=left,                    
    numbersep=5pt,                  
    showspaces=false,                
    showstringspaces=false,
    showtabs=false,                  
    tabsize=2
}

\AtBeginDocument{%
  \providecommand\BibTeX{{%
    \normalfont B\kern-0.5em{\scshape i\kern-0.25em b}\kern-0.8em\TeX}}}

\copyrightyear{2023}
\acmYear{2023}
\setcopyright{acmlicensed}\acmConference[SIGIR '23]{Proceedings of the 46th International ACM SIGIR Conference on Research and Development in Information Retrieval}{July 23--27, 2023}{Taipei, Taiwan}
\acmBooktitle{Proceedings of the 46th International ACM SIGIR Conference on Research and Development in Information Retrieval (SIGIR '23), July 23--27, 2023, Taipei, Taiwan}
\acmPrice{15.00}
\acmDOI{10.1145/3539618.3591816}
\acmISBN{978-1-4503-9408-6/23/07}

\begin{document}
\title{\System{}: A Scalable Entity Alignment System}


\author{Junyang Wu}
\affiliation{%
  \institution{Zhejiang University}
  \city{Hangzhou}
  \country{China}}
\email{wujunyang@zju.edu.cn}


\author{Tianyi Li}
\affiliation{%
  \institution{Aalborg University}
  \city{Aalborg}
  \country{Denmark}}
\email{tianyi@cs.aau.dk}

\author{Lu Chen}
\affiliation{%
  \institution{Zhejiang University}
  \city{Hangzhou}
  \country{China}}
\email{luchen@zju.edu.cn}
\author{Yunjun Gao}
\affiliation{%
  \institution{Zhejiang University}
  \city{Hangzhou}
  \country{China}}
\email{gaoyj@zju.edu.cn}

\author{Ziheng Wei}
\affiliation{%
  \institution{Huawei}
  \city{Hangzhou}
  \country{China}}
\email{ziheng.wei@huawei.com}




\renewcommand{\shortauthors}{Junyang Wu et al.}

\begin{abstract}

Entity alignment (EA) aims to find equivalent entities in different knowledge graphs (KGs).
State-of-the-art EA approaches generally use Graph Neural Networks (GNNs) to encode entities. However, most of them train the models and evaluate the results in a full-batch fashion, which prohibits EA from being scalable on large-scale datasets.
To enhance the usability of GNN-based EA models in real-world applications, 
we present \System{}, a scalable entity alignment system that enables to
(i) train large-scale GNNs for EA, (ii) speed up the normalization and the evaluation process, and (iii) report clear results for users to estimate different models and parameter settings. 
\System{} can be run on a computer with merely one graphic card. 
Moreover, \System{} encompasses six 
state-of-the-art EA models and provides access for users to quickly establish and evaluate their own models.
Thus, \System{} allows users to perform EA without being involved in tedious implementations, such as negative sampling and GPU-accelerated evaluation. With \System{}, users can gain a clear view of the model performance. 
In the demonstration, we show that \System{} is user-friendly and is of high scalability even on computers with limited computational resources.
\end{abstract}

\ccsdesc[500]{Computing methodologies~Knowledge representation and reasoning}
\ccsdesc[300]{Computing methodologies~Semantic networks}

\keywords{Entity Alignment, Knowledge Graphs}


\maketitle

\section{Introduction}
\label{sec:intro}

Knowledge graphs (KGs) represent collections of relations between real-world entities, which facilitates a wide range of downstream applications, 
However, KGs constructed from different sources are highly incomplete~\cite{OpenEA2020VLDB, clusterEA, DualMatch}. Entity alignment (EA)~\cite{OpenEA2020VLDB} has been proposed as a fundamental strategy to complete KGs. Essentially, EA aligns entities from different KGs that refer to the same real-world objects, thereby enabling the completion of KGs.
Graph Neural Networks (GNNs) have attracted significant attention in recent years and GNN-based entity alignment (EA) has emerged as a promising approach to address the problem of KG incompleteness~\cite{GCN-Align18, GAT18, MuGNN19, MRAEA20, RREA20, DualAMN21}.
Despite the advances, there remain two main challenges to be tackled. 

\emph{The first challenge is how to scale up EA models to process large-scale datasets.} The state-of-the-arts~\cite{DualAMN21, RREA20} focus primarily on model design of GNNs, with little attention given to model scalability. Recent open-source EA toolkits (e.g., OpenEA~\cite{OpenEA2020VLDB} and EAKit~\cite{EAKit}) have not yet provided satisfactory solutions for processing large-scale EA. However, the size of real-world KGs is much larger than that of traditional datasets used in evaluating EA models. For instance, a real-world KG YAGO3 includes 17 million entities~\cite{LargeEA22}. 

\emph{The second challenge is how to deal with geometric problems effectively.}  As the size of input KGs increases, some entities tend to frequently appear as the top-1 nearest neighbors of many other entities, while outliers become isolated from the rest of the graph~\cite{clusterEA}. These issues are referred to as geometric problems~\cite{clusterEA, OpenEA2020VLDB} and they can make it difficult to align entities. Although some normalization methods have been proposed to address geometric problems, they are quadratic-time method and thus may still not be feasible for very large graphs. 


To address the limitations of the existing solutions and toolkits, we develop \System{} -- a tool aimed at facilitating the design, development, and evaluation of EA models.  The key contributions of \System{} are summarized as follows.

\noindent
\textbf{Large-scale Training.}
We design a novel solution to enhance the scalability of existing GNN-based.
\System{} first constructs mini-batches based on seed alignment, then it applies neighborhood sampling~\cite{GraphSAGE17} on KGs to generate subgraphes, which can be fed into most GNN-based EA methods. In this way, \System{} allows users to train large-scale graphs for EA on a single GPU. 

\noindent
\textbf{Large-scale Inference and Evulation.}
\System{} introduces an inference module to tackle the geometric problem for large-scale EA. 
First, the system partitions the entities set into several groups with high entity equivalence rates and calculates the local similarity matrix for each group. Next, it normalizes the local matrices and aggregates them into a global similarity matrix. Moreover, we use FAISS~\cite{JDH17} to create sparse matrices using top-$k$ correspondence to speed up the evaluation process.


\noindent
\textbf{The Demonstration of \System{}.}
\System{} offers a suite of visualization tools to help users establish EA tasks and gain insights into the alignment process. It integrates six GNN-based models, including both the classic models~\cite{GCN-Align18, GAT18, MuGNN19} and the state-of-the-art models~\cite{MRAEA20, RREA20, DualAMN21}, making it easy for users to establish EA tasks. In addition, users can customize their GNN models flexibly using Pytorch and Deep Graph Library (DGL)~\cite{DGL19}. The intuitive visualization interface of \System{}
allows users to select models and datasets, explore performance at the entity level, and compare different settings. We have created a demonstration video that can be accessed at https://github.com/Immortals88/Demo-SEA.

\begin{figure}[t]
    \vspace{-4mm}
    \centerline{\includegraphics[width=3.2in]{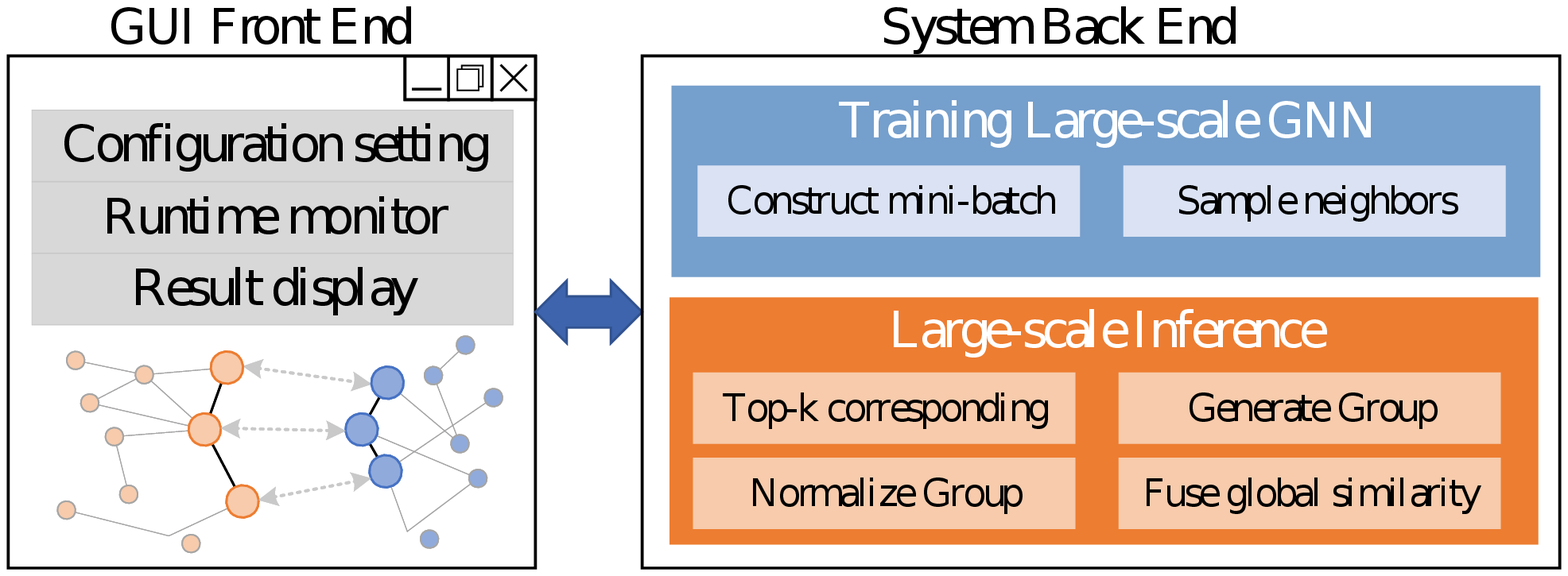}}
    \vspace{-3mm}
    \caption{Architecture of \System{}.}
    \label{overview}
    \vspace{-4mm}
    \end{figure}
\section{System Overview}
\label{sec:define}

Figure~\ref{overview} shows the architecture of \System{}. \System{} consists of a front end with a GUI for users and a back end that manages the EA process. The front end of the system allows users to configure EA settings, monitor the entire training process, and visualize and inspect the final EA results. The back end of the system is composed of two modules responsible for the training, inference, and evaluation of the EA process. These modules collaborate to facilitate EA  across different knowledge graphs. To ease the understanding of \System{}, we first introduce important preliminaries related to the entity alignment process, and then we present a detailed workflow that illustrates how users can utilize the system to perform EA.

\subsection{Preliminaries}
\noindent
 A \textbf{knowledge graph} (KG) is denoted as $G = (E,R,T)$, where $E$ is the set of entities , $R$ is the set of relations, and $T=\{(h,r,t)~|~h,t \in E, r \in R\}$ is the set of triples, each of which represents an edge between the head entity $h$ to the tail entity $t$ with the relation $r$.

\noindent
\textbf{Entity alignment} (EA) aims to find the 1-to-1 mapping of entities $\phi$ from a source KG $G_s = (E_s,R_s,T_s)$ to a target KG $G_t = (E_t,R_t,T_t)$. $\phi = \{(e_s, e_t) \in E_s \times E_t~|~e_s \equiv e_t\}$, where
$e_s \in E_s$, $e_t \in E_t$, and $\equiv$ is an equivalence relation between $e_s$ and $e_t$. In most cases, a small set of equivalent entities $\phi^{\prime} \subset \phi$ is known beforehand and is used as seed alignment.

\noindent
\textbf{GNN-based EA} takes two KGs: $G_s$, $G_t$ and the seed alignment $\phi^{\prime}$ as input, and it uses graph neural networks (GNN) to (i) learn a set of embeddings for all entities $E_s$ and $E_t$, denoted as $\mathbf{f}\in \mathcal{R}^{(|E_s|+|E_t|)\times D}$, where $D$ is the dimension of embedding vectors, and (ii) maximize the similarity (e.g. cosine similarity) of entities that are equivalent in $\phi^{\prime}$.


\begin{figure}[t]

    \vspace{-4mm}
    \centering
    \includegraphics[width=3.2in]{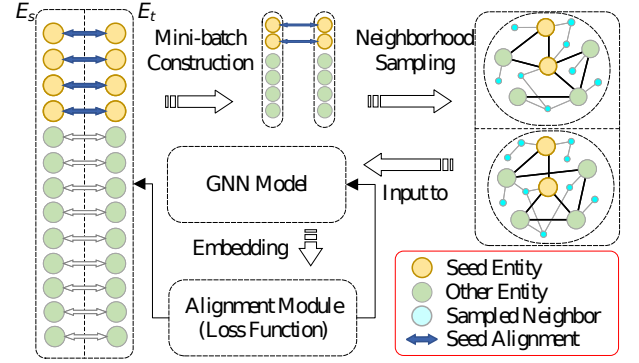}
    \vspace{-3mm}
    \caption{Training large-scale GNNs.}
    \vspace*{-5mm} 
    \label{fig:example}
    \end{figure}
    
\subsection{Training large-scale GNNs for EA}
\label{sec:mini-batch-training}

GNN-based EA approaches have dominated the EA tasks with promising performances by propagating the information of seed alignments to their neighbors. Formally, the embedding $h_{v}^{(k)}$ of an entity $v \in E$ in the $k_{th}$ layer of GNN is obtained by aggregating localized information via $a_{v}^{(k)}=\operatorname{Aggregate}^{(k)}(\{h_{u}^{(k-1)} \mid u \in \mathcal{N}(v) \})$ and
$h_{v}^{(k)}=\operatorname{Update}^{(k)}(a_{v}^{(k)}, h_{v}^{(k-1)})$. 
Where $h_v^{(0)} \in \mathcal{R}^D$ is a learnable embedding vector initialized randomly or generated with some language models.
$\mathcal{N}(v)$ represents the set of neighboring entities around $v$. The model's final output on entity $e$ is denoted as $\mathbf{f}_e$.

Most GNN-based EA approaches train their models in a whole graph manner. It is no longer feasible when the KGs become large. 
 Since the storage of the entity embeddings and graph structure may exceed the memory capacity.
To address this issue, \System{} provides a general workflow for training a Siamese GNN on large graphs: (i) constructs mini-batch of entities; and (ii) samples neighborhoods in each mini-batch as a subgraph for GNN information propagation as depicted in Figures~\ref{fig:example}.

\noindent
\textbf{Mini-batch Construction.}
To scale up the existing GNN-based EA models, \System{} trains these models with neighborhood sampling strategy~\cite{GraphSAGE17}. 
Before we start to sample subgraph, we need to construct a bunch of entities, called mini-batch, as starting points to kick off the sampling process. Each mini-batch should keep the same setting as the whole entity set. Specifically, \System{} first randomly selects a set of entities from the seed alignment that are equivalent. Next, it picks negative nodes from the entire entity set that do not overlap with seed entities. Finally, a mini-batch consisting of these two parts (positive and negative entities) is generated as shown in Figures~\ref{fig:example}.


\begin{figure*}[t]
    \vspace{-4mm}
    \centering
    \includegraphics[width=6.8in]{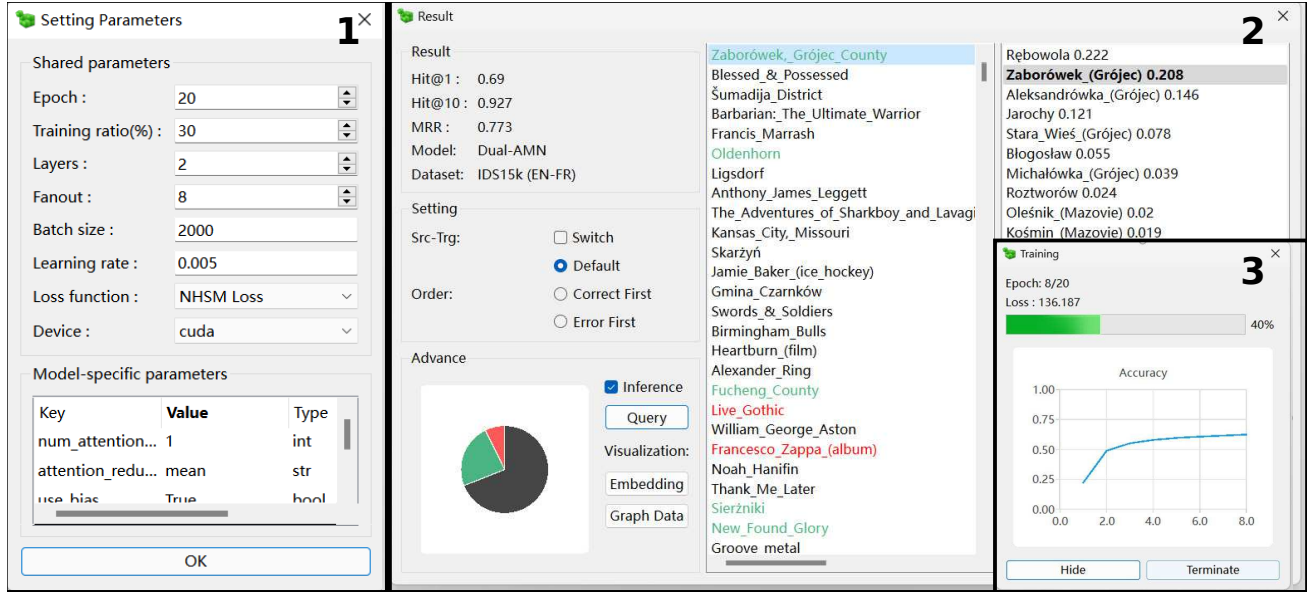}
    \vspace{-3mm}
    \caption{The Graphic User Interface of \System{}.}
    \label{fig:demo_gui}
    \vspace*{-4mm}
    \end{figure*}

\noindent
\textbf{Neighborhood Sampling.} 

In \System{}, the entire graph information is stored in CPU memory. When a mini-batch is generated, \System{} samples the $k$-hop neighborhoods for each entity in the mini-batch to form a subgraph. The hop number $k$ is determined by the number of layers defined in GNN models. By controlling the size of the mini-batch, the hop number, and the fan-out (i.e. maximum sampled neighbors) for each node, the resulting subgraph is guaranteed to be small enough to fit in GPU memory. The subgraph and the embeddings of the involved entities are then shifted to the GPU for GNN training. This significantly reduces the GPU memory cost, making it possible to train large-scale inputs (e.g., DBP1M~\cite{LargeEA22}).


\noindent
\textbf{Model Training and Loss Functions.}
The GNN models used in \System{} take the sampled subgraphs as input and generates the embedding of the entities in the mini-batch as output. The loss function is designed to encourage the embeddings of equivalent entities to be close together while keeping non-equivalent entities far apart. This allows us to measure the distance between entities and identify potential alignments.
In \System{}, we implement multiple GNN models (cf. Section \ref{model and data}) and two commonly used loss functions: the triplet loss~\cite{GCN-Align18} and the normalized hard sample mining loss~\cite{DualAMN21}. 
Note that the users can define their own loss functions. 

\subsection{Large-scale Inference}
\label{inference}
After obtaining the KG embeddings, we can get the similarity matrix by measuring the distances between them. However, as the size of KGs grows, the geometric problems become more severe~\cite{clusterEA}. Many existing studies~\cite{clusterEA, OpenEA2020VLDB, DualMatch} use normalization methods on the similarity matrix to alleviate the geometric problems. But they have at least quadratic time complexity.
To solve the problem, \System{} first groups entities that have a high likelihood of being equivalent together, and computes the local similarity matrix within each group. Next, \System{} uses FAISS with GPU acceleration to calculate the top-$k$ correspondences, which are then used to form a global sparse similarity matrix. Finally, \System{} normalizes the local similarity matrices and merges them with the global sparse similarity matrix to find the alignment, greatly reducing the time complexity.

Maintaining performance while reducing complexity poses two challenges: (i) generating groups with a high entity equivalence rate to ensure a 1-to-1 mapping, and (ii) merging the local similarity matrices to ensure overall accuracy. To address these challenges, in \System{}'s inference stage, we adopt our previous work, \Cluster{}~\cite{clusterEA}. \Cluster{} trains a classifier to group entities with a high equivalent rate and proposes a SparseFusion method to normalize the global similarity matrix effectively and efficiently.

\System{} offers two types of evaluation metrics to assess the performance of EA: (i) the Hits-$k$ metric (H@$k$), indicating whether the ground truth appears in top-$k$ ranking; and (ii) the Mean reciprocal rank (MRR), indicating the average of the reciprocal rank of the ground truth. 

\subsection{Models and Dataset}
\label{model and data}

\noindent
\textbf{Models.}
In \System{}, we implement the stochastic training of six GNN-based EA models using DGL~\cite{DGL19}. These models are 
(i) \emph{GCNAlign}~\cite{GCN-Align18}, the first GNN-based EA model; 
(ii) \emph{GAT}~\cite{GAT18}, a common GNN model that introduces attention technique and has been used as a building block in many recent approaches ~\cite{DualAMN21, DualMatch};
(iii) \emph{MRAEA}~\cite{MRAEA20}, a GNN-based EA model that models entity embeddings by integrating the node’s incoming and outgoing neighbors and the connected relations’ meta semantics;
(iv) \emph{MuGNN}~\cite{MuGNN19}, a GNN-based EA model that utilizes multi-channel information to align entities across different knowledge graphs; 
(v) \emph{RREA}~\cite{RREA20}, a GNN-based EA model that leverages relational reflection transformation to obtain relation-specific embeddings for each entity; 
and (vi) \emph{Dual-AMN}~\cite{DualAMN21}, an EA model that encompasses two attention layers to model both intra-graph and cross-graph relations.

\noindent
\textbf{Datasets.}
\System{}  has built-in support for five commonly used real-world datasets:  DBP15K~\cite{Websoft},
DWY100K~\cite{Websoft}, 
SRPRS~\cite{Websoft}, 
IDS~\cite{Websoft},
and a large-scale dataset DBP1M~\cite{LargeEA}.


\section{Demonstration}



\System{} is a cross-platform application that utilizes PyTorch for its back-end logic and Qt for its graphic user interface. Its basic interface, shown in Figure~\ref{fig:demo_gui}, includes functionalities such as hyperparameter configuration, EA model training, alignment results evaluation, and visualization. This demonstration serves to showcase the \System{}.

\noindent
%
\noindent
\textbf{Hyper-parameter Configuration.} \System{} offers support for six real-world datasets commonly used in EA (cf. Section~\ref{model and data}). Users can either choose one of the supported datasets or load a local dataset. Then, \System{} checks the input format for compatibility upon loading. Once the dataset is loaded, users are required to select an EA model for training and to set the corresponding hyper-parameters through interface No.1, as shown in Figure~\ref{fig:demo_gui}. To simplify the process, some suggested hyperparameters are pre-filled based on the selected model and dataset, eliminating the need for non-experts to manually tune them.

The hyper-parameters in \System{} are divided into two parts: (i) \textit{Shared hyperparameters}, which must be specified by all EA models, such as learning rate, loss functions, model layer number, and training epochs; and (ii) \textit{Model-specific parameters}, which must be set specifically for the selected model, such as the number of attention heads in Graph Attention Network (GAT) related approaches.


\begin{figure}[t]
\centering
\begin{lstlisting}[language=Python]
class SimpleGCN(nn.Module):
    def __init__(self, in_feats, out_feats):
        super(SimpleGCN, self).__init__()
        self.conv = dgl.nn.GraphConv(in_feats, out_feats, 
            norm='none', weight=True, bias=True)
        xavier_normal_(self.conv.weight)
        zeros_(self.conv.bias)
    def forward(self, blocks):
        in_feats = blocks[0].srcdata['feature']
        edge_weight = blocks[0].edata['weight']
        h = self.conv(blocks[0], in_feats,
            edge_weight=edge_weight)
        return h
\end{lstlisting}
\vspace{-3mm}
\caption{An example code of implementing a one-layer GCN for the EA task in \System{}.}
\vspace*{-4mm}
\label{code:simple_gcn}
\end{figure}

\noindent
\textbf{User-defined GNN models.}
\System{} provides the ability to implement new GNN models for EA. Users can define their own GNN models, as depicted in Figure~\ref{code:simple_gcn}. To define a new model, a user only needs to follow the guideline of PyTorch and DGL~\cite{DGL19} to inherit the \textit{torch.nn.Module} class and implement two methods: \textit{\_\_init\_\_()} that constructs the model and \textit{forward(blocks)} that takes a list of sampled subgraph as input and outputs the node embedding of its destination nodes. The input of the {\it forward} function is a series of \textit{DGLBlocks}~\cite{DGLToBlock} representing the sampled subgraph, where each one yields the node feature and edge weights of its layer. The model design is similar to building a model for node classification tasks, which simplifies the task for users by eliminating the need to implement tedious procedures.

\begin{figure}[t]
    \vspace{-1mm}
    \centering
    \subfigtopskip=-5pt
    \subfigcapskip=-5pt
    \subfigure[graph structure]{
        \label{subfig:neo}
        \includegraphics[width=2.2in]
        {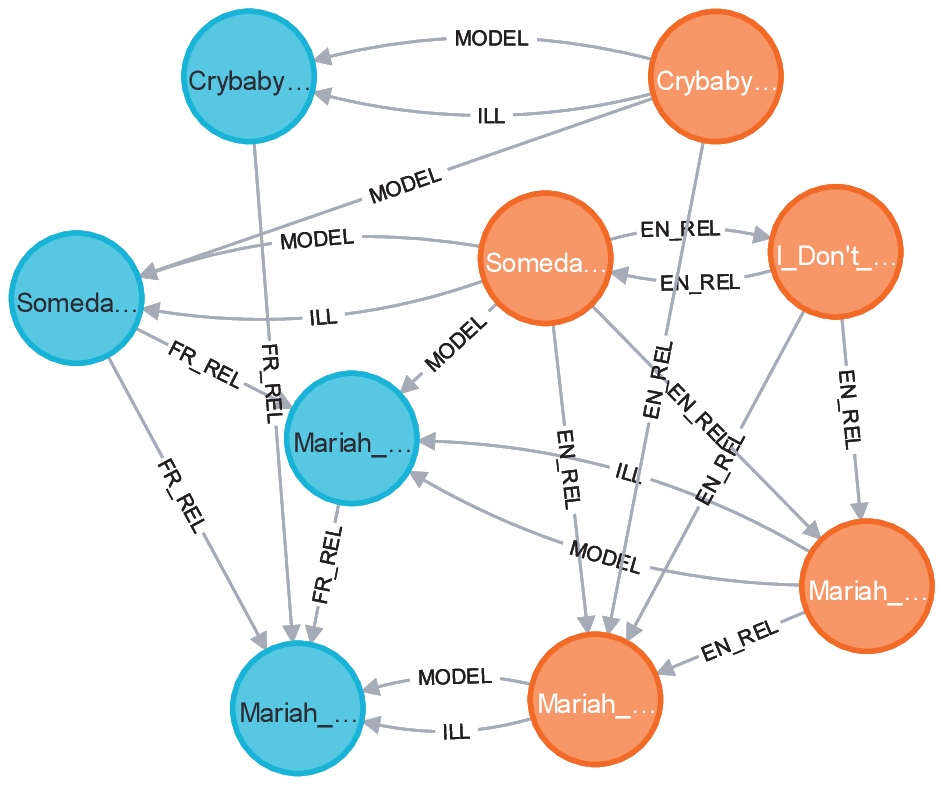}
    }
    \subfigure[train pairs]{
        \label{subfig:train_emb}
        \includegraphics[width=1.55in]{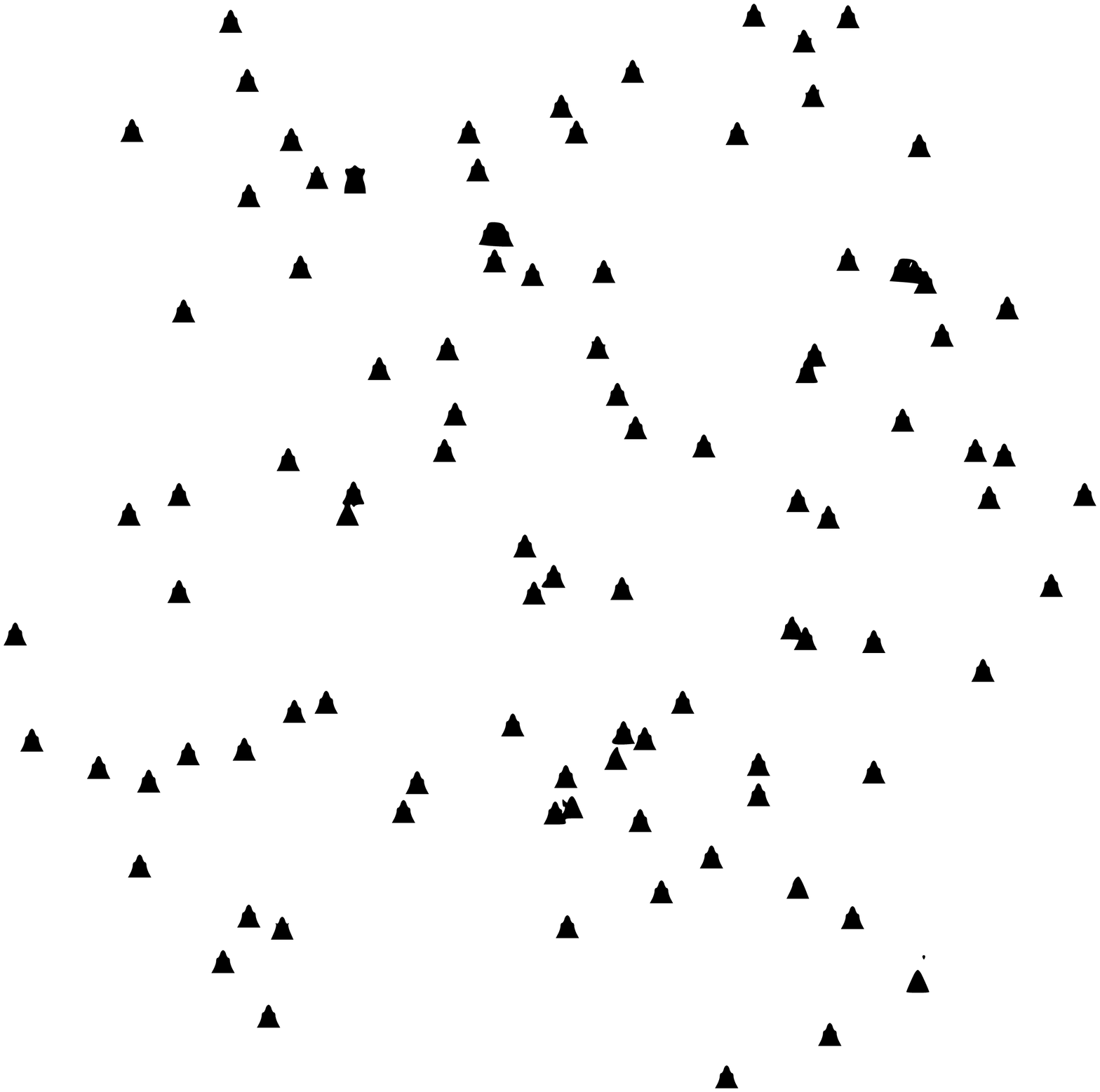}
    }
    \subfigure[test pairs]{
        \label{subfig:test_emb}
        \includegraphics[width=1.55in]{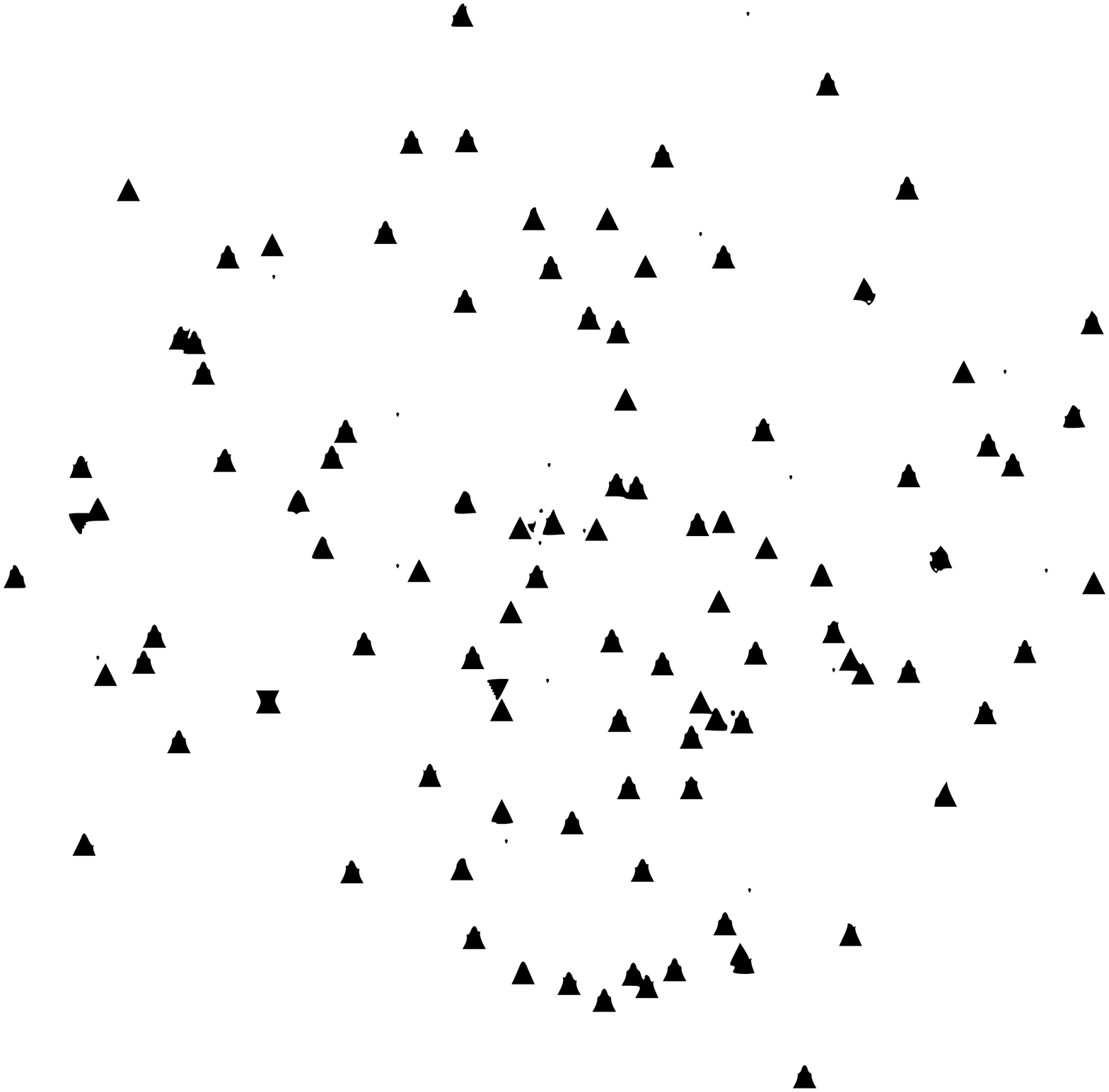}
    }
    \vspace{-2mm}
    \caption{Graph data visualization.}
    \vspace{-5mm}
    \label{fig:demo_g}
    \end{figure}

\noindent
\textbf{Training of EA model.} Once the parameters has been set, the user can initiate the EA task by clicking the "Run!" button. The training process, as outlined in Section~\ref{sec:mini-batch-training}, is fully parallelized on the GPU. \System{} offers real-time tracking of the training loss and prediction accuracy(cf. No.2 in Figure~\ref{fig:demo_gui}). Users can further enhance tracking by integrating other visualization tools such as TensorBoard.

\noindent
\textbf{Results Evaluation \& Visualization.} When the EA process is completed, the system proceeds to the alignment results visualization stage, which contains two parts: \emph{align result visualization} and \emph{graph data visualization}, as shown in Figures~\ref{fig:demo_gui} and~\ref{fig:demo_g}, respectively. 


For the \emph{align result visualization}, we provide basic evaluated information including \HitOne{}, \HitTen{} and \MRR{} generated by our large-scale evaluation unit. To facilitate the interpretability of EA models, \System{} depicts the EA results in the form of lists. Entities in set $E_s$ are shown in a list on the left-hand side. If we click one entity in the list, \System{} will show the top-10 corresponding entities in $E_t$ produced by the EA model, with the ground truth highlighted (if any). If the entity is properly aligned, it is shown in black. If the top-10 correspondences include the ground truth (but not as the top choice), the entity is shown in green. If the ground truth is not found in the top-10 correspondences, the entity is shown in red.
Several options are provided to re-arrange the list as depicted in Figure~\ref{fig:demo_gui}, No.2. In particular, users can switch the source and target and prioritize the erroneous entities. Moreover, \System{} maintains a class for each entity, which includes the entity's name, its neighbors, the ground truth, and a list of candidate entities generated by the model. This means users can write their own queries to select specific entities they wish to examine. As described in Section~\ref{inference}, we have proposed an inference method to address geometric problems. By clicking the "inference" checkbox, users can observe the effectiveness of this module. If the box is not checked, \System{} will identify the nearest neighbor for each entity in a naive way. This leads to a drop of 20\% in Hits@1 for almost all settings.

For the \emph{graph data visualization}, \System{} provides visualization tools for both KG structure and entity embeddings.
To depict the two KGs and their alignment, we have integrated \System{} with Neo4j, as shown in Figure~\ref{subfig:neo}. The nodes of the two graphs are painted with different colors (orange for EN and blue for FR in this example). The top-10 alignment produced by the EA model is added as edges (MODEL) between graphs, along with the ground truth edges (ILL). Users can utilize Cypher Query Language to explore the aligned result in more detail.  
As for embedding visualization, we use a dimension reduction method TSNE to project embedding vectors to 2D points, where the distance between points reflects their distance in the high-dimensional space.  Figure~\ref{subfig:train_emb} shows that the embeddings of the training data are highly aligned, while the test pairs have some minor errors. By exploring the embedding visualization, users can gain more insight into the EA process.
    
\begin{acks}
This work was supported in part by the National Key Research and Development Program of China under Grant No. 2021YFC3300303, the NSFC under Grants No. (62025206, 61972338, and 62102351). Tianyi Li is the corresponding author of the work.
\end{acks}


\bibliographystyle{ACM-Reference-Format}
\bibliography{refer}


\end{document}